\documentclass[conference]{IEEEtran}
\IEEEoverridecommandlockouts
\usepackage{cite}
\usepackage{amsmath,amssymb,amsfonts}
\usepackage{algorithmic}
\usepackage{graphicx}
\usepackage{textcomp}
\usepackage{hyperref}
\usepackage{xcolor}
\usepackage{soul}
\usepackage{subcaption}

\def\BibTeX{{\rm B\kern-.05em{\sc i\kern-.025em b}\kern-.08em
    T\kern-.1667em\lower.7ex\hbox{E}\kern-.125emX}}
\begin{document}

\title{Improved Generation of Synthetic Imaging Data Using Feature-Aligned Diffusion}

\author{\IEEEauthorblockN{Lakshmi Nair}
\IEEEauthorblockA{\textit{Researcher} \\
Boston, Massachusetts, USA \\}
}

\maketitle

\begin{abstract}
Synthetic data generation is an important application of machine learning in the field of medical imaging. While existing approaches have successfully applied fine-tuned diffusion models for synthesizing medical images, we explore potential improvements to this pipeline through \textit{feature-aligned diffusion}. Our approach aligns intermediate features of the diffusion model to the output features of an expert, and our preliminary findings show an improvement of 9\% in generation accuracy and $\approx0.12$ in SSIM diversity. Our approach is also synergistic with existing methods, and easily integrated into diffusion training pipelines for improvements.  We make our code available at \url{https://github.com/lnairGT/Feature-Aligned-Diffusion}
\end{abstract}


\section{Introduction}
The field of healthcare has seen transformative changes following the recent advancements in generative AI, from protein folding \cite{abramson2024accurate}, to foundational models for genomics data \cite{cui2024scgpt}. Among the different applications of machine learning in healthcare, synthetic data generation is an important area of focus in order to address privacy concerns while decreasing costs \cite{giuffre2023harnessing}. Additionally, synthetic data helps supplement limited training data availability that is common in healthcare applications, enabling the training of larger and better models. Particularly, in the medical imaging domain, data scarcity is a common problem caused due to factors such as expensive image acquisition, labeling procedures, privacy concerns and rare incidences of certain pathologies \cite{montoya2024mam}.

\section{Related Work and Motivation}
Recent approaches to synthetic data generation leverages state-of-the-art performances of diffusion models. In \cite{kidder2023advanced}, the authors use a fine-tuned stable diffusion model with DreamBooth, for the synthesis of MRI scans. DreamBooth \cite{ruiz2023dreambooth}, uses a few images of a new subject with a respective, unique text identifier to fine-tune the diffusion model. Similarly, \cite{farooq2024derm} use Stable diffusion and DreamBooth for generating synthetic images of skin lesions. In order to evaluate the synthetic generations, the authors use two state-of-the-art skin lesion classifiers: ViT and Mobilenet-v2. For synthetic generation of mammograms, authors in \cite{montoya2024mam} introduce a two-part approach: one model for healthy mammogram generation and another for lesion in-painting. In \cite{khader2023denoising}, the authors demonstrate synthetic generation of MRI and CT scans by applying fine-tuned diffusion models, evaluating the synthesized images with the help of expert radiologists. In contrast to diffusion models, prior work has also explored the use of GANs, particularly StyleGAN2, in the generation of synthetic data \cite{ding2023large}. While effective, the use of GAN-like architectures in medical image synthesis is challenging due to unstable training, low sample diversity and quality \cite{kazerouni2023diffusion}.

\begin{figure}
\centering
  \includegraphics[width=0.47\textwidth]{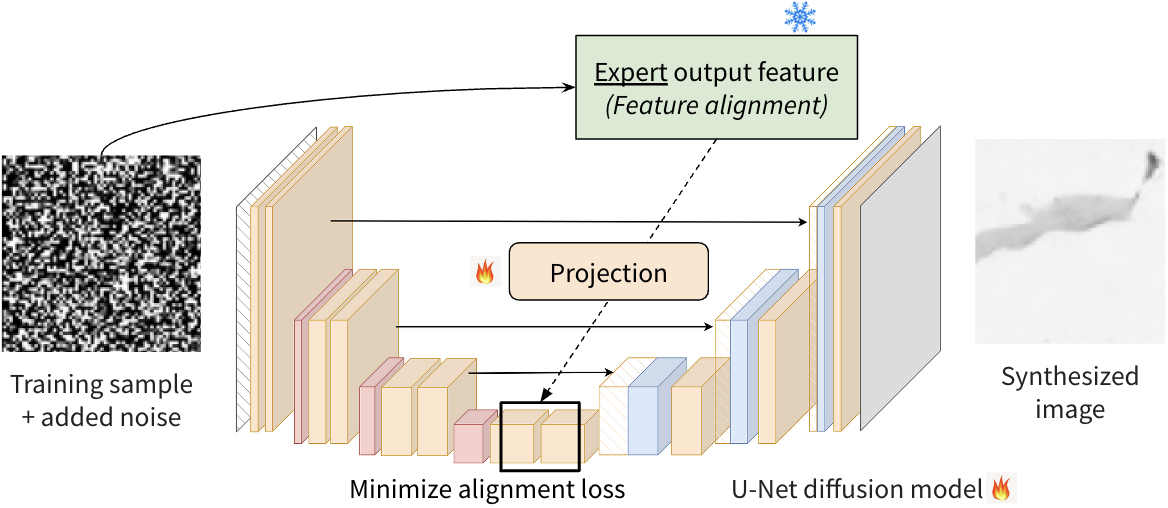}
  \caption{Overview of feature-aligned training of diffusion models. Example shows synthesis of Adipose tissue image.}
  \label{fig:overview}
\end{figure}

All the above approaches involve fine-tuning existing diffusion models directly, i.e., the relevant training data is pre-processed, and a diffusion model is fine-tuned on the desired data, either with or without DreamBooth. In this brief paper, \textbf{\ul{we explore whether aligning the diffusion model with features extracted from an \textit{expert model}, can help improve the quality of generations}}. Hence, one assumption of this work is the existence of an ``expert'' model. However, ``expert'' models are commonly used to evaluate synthetic generations and therefore \textit{already available} in such scenarios \cite{ding2023large,farooq2024derm}. Hence, \ul{our approach is complementary to existing methods to further augment their generation accuracy, and is easily incorporated into existing diffusion training pipelines with only an additional loss term and projection layer}.

Our key finding is that aligning the intermediate features of the diffusion model with output features of a classification expert during training, can lead to improved generations during inference. Interestingly, we observe that these improvements occur when the expert features are computed on the \textit{noise added inputs} that are fed to the diffusion model during training, as opposed to the noise free original training samples. In other words, aligning with the expert features of the \textit{noisy image} as opposed to the noise-free image, leads to improvements during inference.


\section{Diffusion -- Preliminaries}
\label{subsec:diff_prelim}
Diffusion models are latent-variable generative models that generate data by iteratively de-noising a sample from Gaussian noise \cite{ho2020denoising}. The diffusion model formulation consists of a fixed forward process, that takes a data sample from an initial distribution, progressively corrupting it with Gaussian noise. It also consists of a reverse process that learns to undo this corruption, effectively recovering samples from the original data distribution.

The forward process transforms a data point into a noisy version over discrete timesteps. At each timestep, noise is incrementally added according to a predefined variance schedule $\alpha_t$. In terms of samples $x_t$, this process can be formulated as:
\begin{equation}
    x_t = \alpha_t x_0 + (1 - \alpha_t) \epsilon, \text{  } \epsilon \sim \mathcal{N}(0, 1)
\end{equation}
The reverse process involves learning a model that predicts the noise added at each step, directly recovering an estimate of $x_0$. This is typically done with a loss function as follows:
\begin{equation}
    L_{noise} = \|\epsilon - \epsilon_\theta(x_t, t)\|^2_2
\end{equation}
The generations can be conditioned on class labels, where the model additionally takes the desired class label in the form of their corresponding text embeddings (obtained with a text encoder). It is then incorporated into the diffusion model via cross-attention layers \cite{rombach2021high}, allowing the generations to be controlled by the specific label.

\subsubsection{Diffusion Model Architecture}
U-Net architectures are a common choice for diffusion models that demonstrate state-of-the-art performances \cite{rombach2021high}. A U-Net consists of a downsampling block (contracting path) and an upsampling block (expanding path), with residual connections between the two paths. The downsampling block consists of a series of layers that gradually reduce spatial information while capturing \textit{feature} information. In contrast, the upsampling block gradually recovers the spatial information from the feature information of the downsampling block.

\section{Feature-Aligned Diffusion}
\label{subsec:feat_align}
We propose feature-aligned diffusion to improve model generations, by aligning intermediate features of the diffusion model with the output features of an expert during fine-tuning. In this context, the expert model refers to a classification model, often used to evaluate synthetic generations \cite{farooq2024derm,ding2023large}. Our approach is shown in Figure \ref{fig:overview}.

Typical diffusion model training involves adding noise according to a pre-specified schedule to each training sample passed into the diffusion model. Here, the loss function (from Section \ref{subsec:diff_prelim}) compares the model predicted noise to the true noise added to the sample. Feature alignment incorporates one additional step into the typical flow: we additionally pass the \textit{noisy} training sample to the expert model, to compute the corresponding output features. Intermediate features of the diffusion model are then extracted and aligned with the expert features. In our case, the intermediate features are obtained from the output of the downsampling block of the diffusion U-Net (shown in Figure \ref{fig:overview}). In order to align the intermediate diffusion model features to the output features of the expert, we introduce the following loss function that maximizes the cosine similarity between the two. Computing cosine similarity in this manner requires the expert feature dimensions $E_e$ to match the intermediate diffusion feature dimensions $E_d$. Hence, we also add an additional trainable projection $W_p$:
\begin{equation}
x'_t = f_{e}(x_t)
\end{equation}
\begin{equation}
    L_{align} = -D_c(W_p \cdot x'_t, \text{  } f_{d}(x_t))
\end{equation}
Here, $D_c$ denotes cosine similarity, $x_t$ is the input image \textit{with added noise} (Eqn 1); $W_p$ denotes the projection layer $\mathbb{R}^{E_e \times E_d}$; $f_{e}(\cdot)$ is the output of the expert model, and $f_{d}(\cdot)$ are intermediate features of the diffusion model, extracted at the output of the downsampling block. When training the feature-aligned diffusion model, we use a weighted sum of $L_{noise}$ (Eqn 2) and $L_{align}$ (Eqn 4) for the combined loss function:
\begin{equation}
    L = w_1 \cdot L_{noise} + w_2 \cdot L_{align}
\end{equation}
We note that the expert feature alignment is \textit{only applied during fine-tuning}, i.e., the expert model is \textit{not} needed during inference. During inference, the diffusion model is provided with a class label and input noise to generate images corresponding to the class.

\textbf{Note on processing of features:} Common in architectures like ResNet50 \cite{he2015deep}, the expert features of dimensions $(B, E_e, H, W)$ are typically passed through an adaptive average pooling layer, resulting in a $(B, E_e)$ output. Here, $B$ is the batch size, and $(H, W)$ denote the spatial dimensions of the image. This pooling is done prior to the output being used for classification (via linear and softmax layers). We extract the output of the adaptive average pooling layer and pass it into the projection $W_p$ to create an output of size $(B, E_d)$. For the diffusion model, the downsampling block similarly yields an output of dimensions $(B, E_d, H', W')$. We pass this into an adaptive average pooling layer to yield a $(B, E_d)$ output. Following this processing, the expert features and intermediate diffusion features can be directly compared via cosine similarity.

\begin{figure}
\centering
  \includegraphics[width=0.475\textwidth]{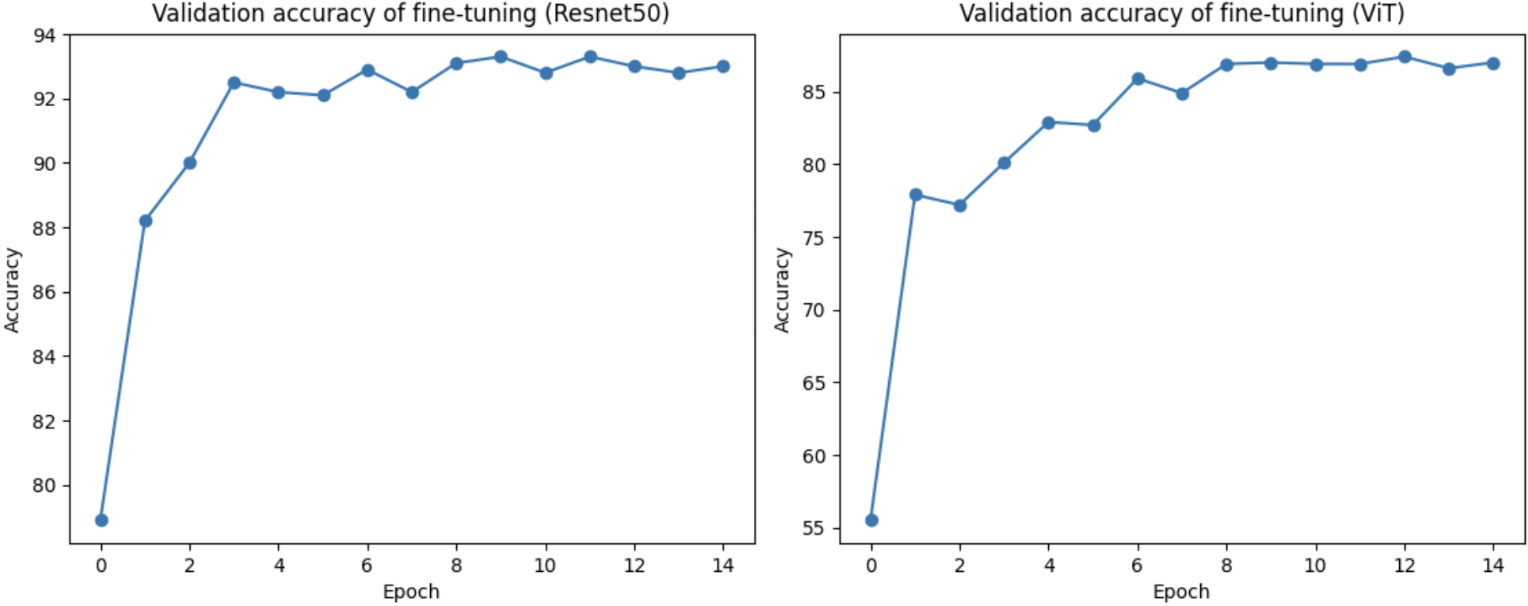}
  \caption{Fine-tuning validation accuracy of expert models -- ResNet50 (left) with 93\% and ViT with 87\%.}
  \label{fig:expert_val}
\end{figure}

\begin{figure}
\centering
  \includegraphics[width=0.47\textwidth]{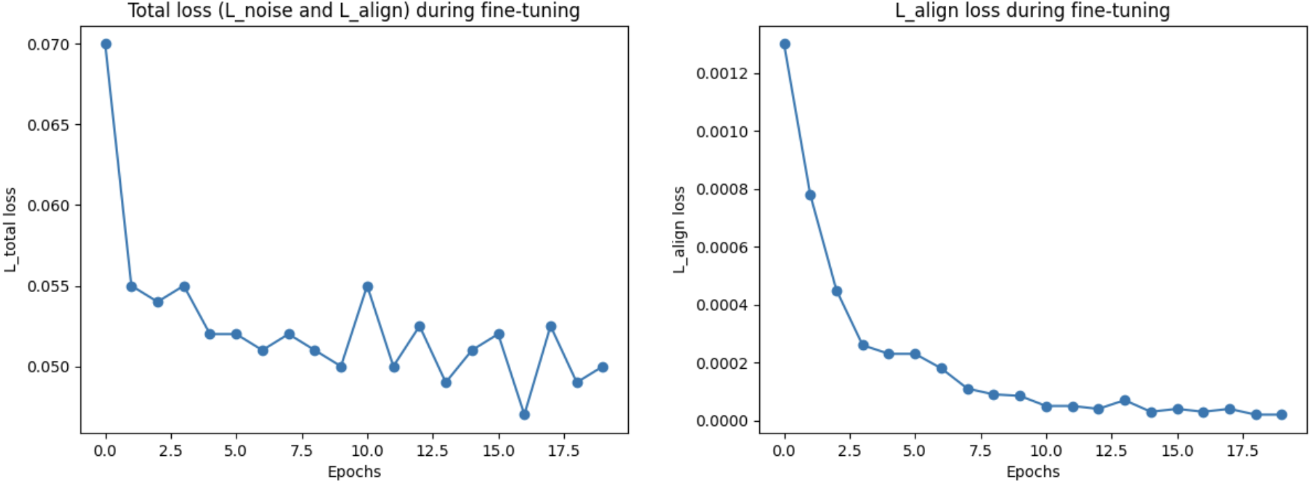}
  \caption{Loss during fine-tuning with feature alignment.}
  \label{fig:loss}
\end{figure}

\begin{table}[]
\centering
\caption{Generation accuracy when aligning to expert features computed on the original noise-free training samples vs. expert features computed on noise-added training inputs.}
\begin{tabular}{c|c}
\textbf{Alignment}   & \textbf{Generation Accuracy} \\ \hline
To noise-free input  & 28.57\%                      \\
To noise-added input & 67.14\%
\end{tabular}
\label{tab:align_inputs}
\end{table}

\textbf{Why would feature-aligned diffusion work?} Our intuition comes from prior work demonstrating that preference optimization is possible within the noisy, latent space of the U-Net model \cite{gambashidze2024aligning}. Their work evaluates the downsampling output at timestep $t-1$: one timestep before the predicted noise, comparing it to the ground truth of applying one reverse step to the true added noise, in order to improve model generations. In contrast to their work, we explore whether a direct alignment between an expert and the diffusion model can be performed within this noisy, latent space.

\section{Experiments}
For our experiments, we use an existing dataset of histological images of colorectal cancer across 8 tissue classes \cite{kather2016multi}. We convert the images from the dataset to gray-scale in this work. For the model architecture, we use an open-source Stable Diffusion \cite{rombach2021high} model from HuggingFace (\textit{segmind/tiny-sd}) consisting of interleaved resnet and cross-attention blocks in the upsampling and downsampling paths. We make our code available at \url{https://github.com/lnairGT/Feature-Aligned-Diffusion}.

\subsection{Choice of Expert Model}
For the expert model, we explored fine-tuning of two different architectures: ResNet50 \cite{he2015deep} and ViT \cite{dosovitskiy2020image} -- both models were pretrained on the ImageNet-1k dataset. The models were fine-tuned with a learning rate of 1e-4, batch size of 64 and image size of 224. We fine-tuned for 15 epochs with a weight decay of 0.7. We found that ResNet50 achieved a classification accuracy of 93\%, whereas ViT achieved 87\% (Figure \ref{fig:expert_val}). Hence, we use ResNet50 as our expert.

\begin{figure}
\centering
  \includegraphics[width=0.38\textwidth]{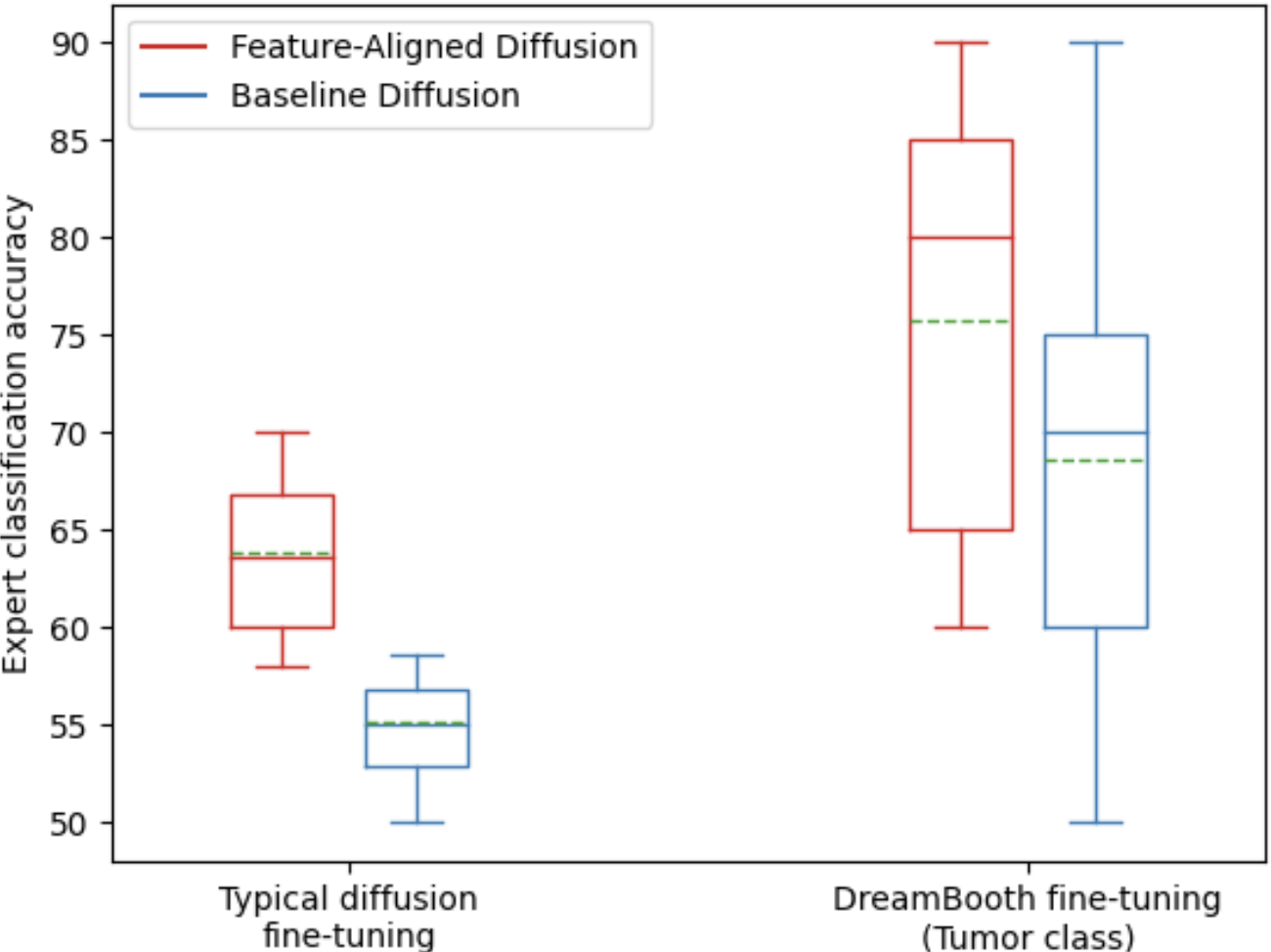}
  \caption{Generation accuracy of feature-aligned vs. baseline diffusion for two fine-tuning pipelines: typical and DreamBooth}
  \label{fig:box}
\end{figure}

\begin{figure}
\centering
\begin{subfigure}[b]{0.36\textwidth}
   \includegraphics[width=1\linewidth]{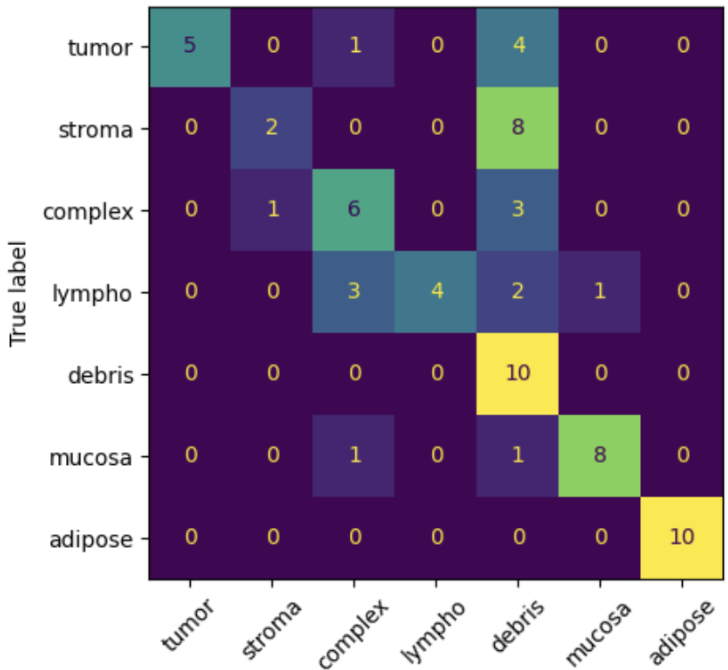}
   \caption{}
   \label{fig:cf_diff} 
\end{subfigure}

\begin{subfigure}[b]{0.36\textwidth}
\centering
   \includegraphics[width=1\linewidth]{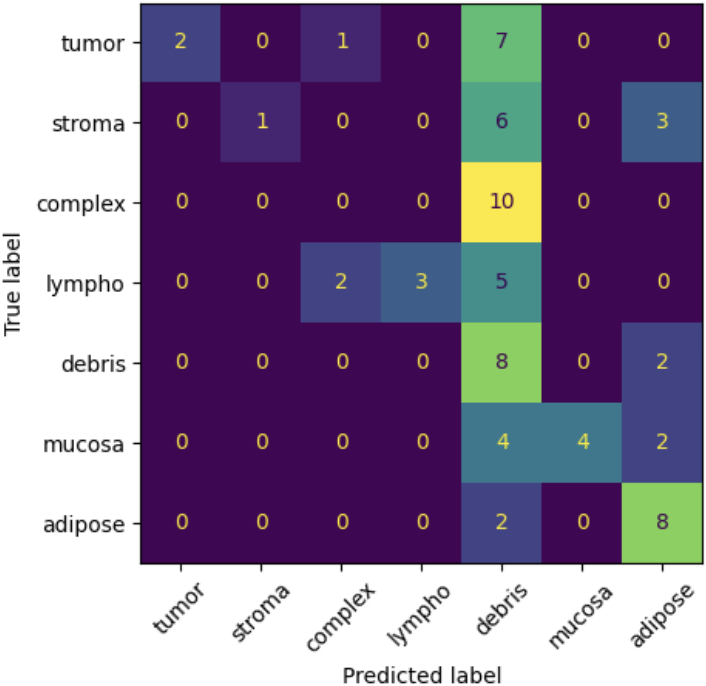}
   \caption{}
   \label{fig:base_cf_mat}
\end{subfigure}

\begin{subfigure}[b]{0.36\textwidth}
   \includegraphics[width=1\linewidth]{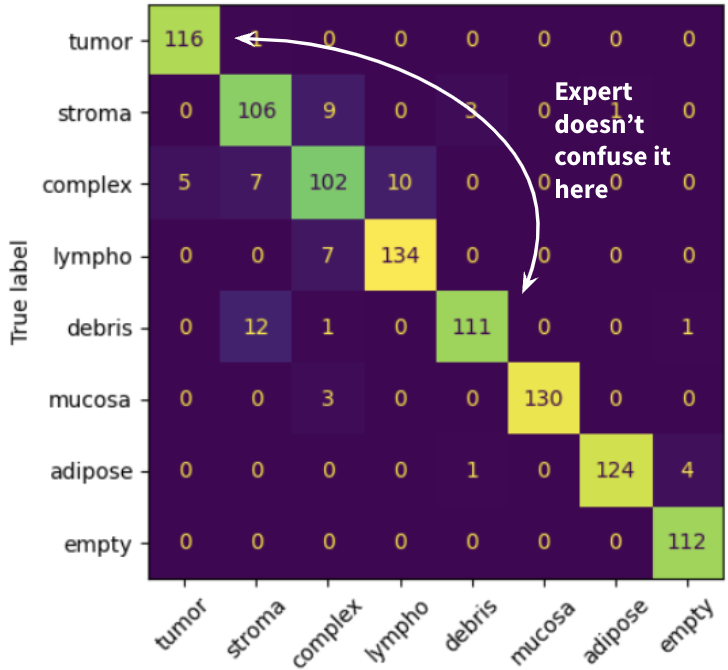}
   \caption{}
   \label{fig:cf_exp}
\end{subfigure}

\caption[Two confusion matrices]{(a) Confusion matrix of expert on feature-aligned diffusion generations, (b) Confusion matrix of expert on baseline diffusion generations (c) Confusion matrix of expert on original dataset. The expert does not mis-classify tumor, stroma, and lympho as debris within the original dataset.}
\end{figure}

\begin{figure}
\centering
  \includegraphics[width=0.465\textwidth]{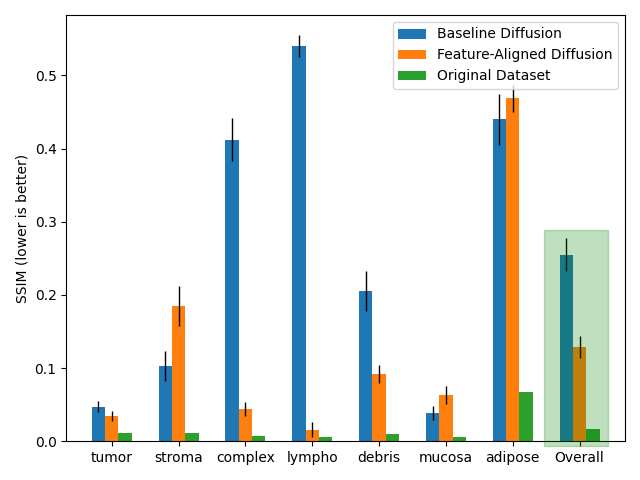}
  \caption{SSIM of baseline, feature-aligned diffusion and original dataset. Note: Lower SSIM $\rightarrow$ better sample diversity.}
  \label{fig:ssim}
\end{figure}

\subsection{Diffusion Fine-tuning Hyper-parameters}
We fine-tune the diffusion models for 20 epochs, either with or without feature alignment, using a learning rate of 1e-4. We use a batch size of 4, and image size of 64. During expert evaluation, the images are resized to 224. For the loss with feature alignment (Eqn 5), we set $w_1 = w_2 = 1.0$. Diffusion timesteps are set to 1000. We use HuggingFace \textit{diffusers}.

Figure \ref{fig:loss} shows the loss curves during feature-aligned fine-tuning, for the combined loss ($L_{noise}$ and $L_{align}$), and for just $L_{align}$. We see the model effectively optimize for $L_{align}$.

\begin{figure*}
\centering
  \includegraphics[width=0.873\textwidth]{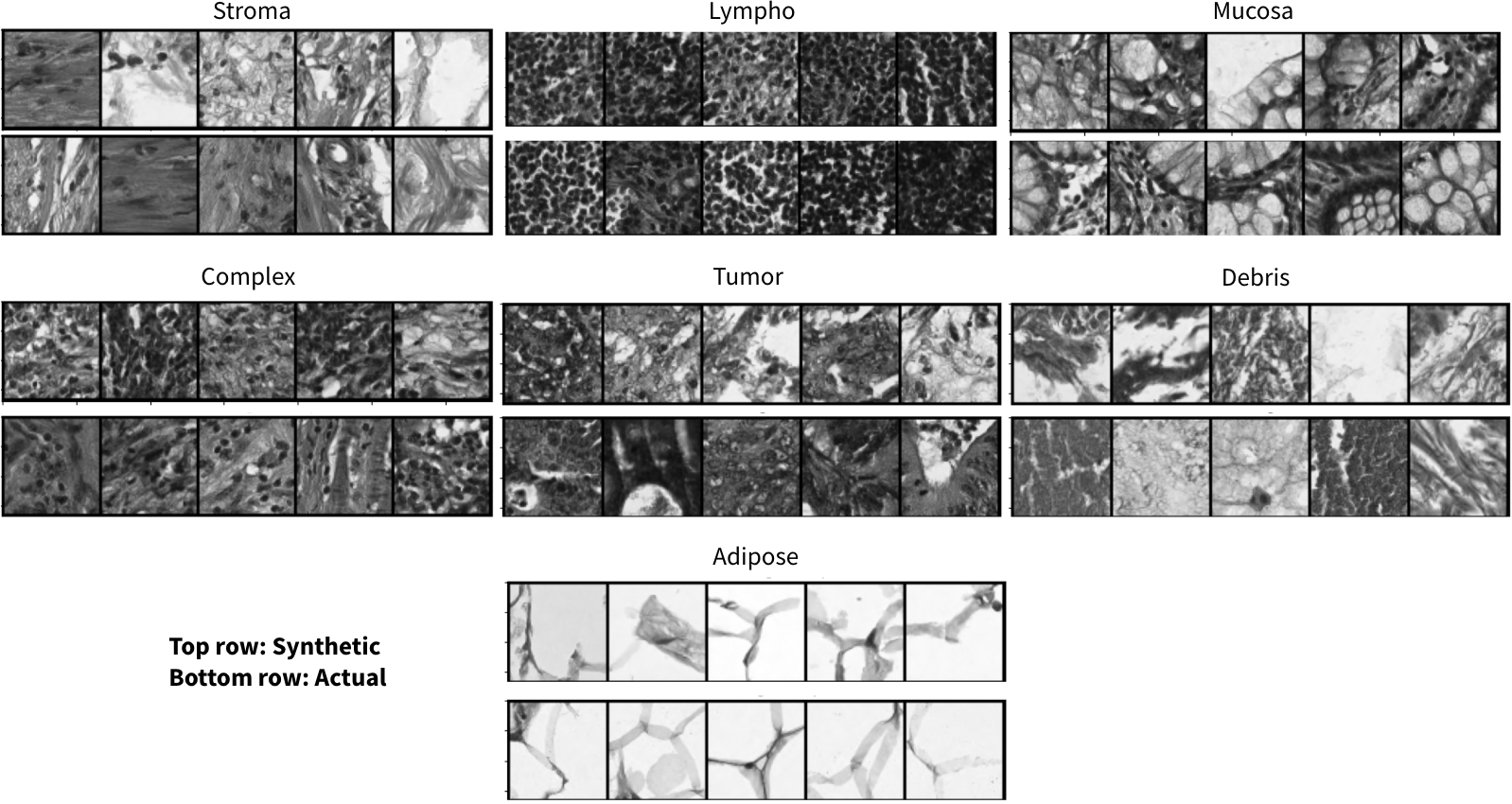}
  \caption{Example generations of the feature-aligned diffusion model.}
  \label{fig:generations}
\end{figure*}

\subsection{Evaluation Method}
We generate 10 synthetic images per class of tissue, excluding the ``empty'' class (since it contains no tissue). We use the expert ResNet50 model to classify the generated images and report the resultant accuracy. We compare performances of the baseline diffusion model (Section \ref{subsec:diff_prelim}) against the feature-aligned diffusion model (Section \ref{subsec:feat_align}) for the two pipelines most commonly used in medical image synthesis \cite{farooq2024derm,kidder2023advanced,montoya2024mam}: a) typical/vanilla fine-tuning, and b) DreamBooth fine-tuning for one class of images (we use the ``Tumor'' class). When comparing the models, we fix the random seeds used for generation to ensure a fair comparison of the model outputs. We note average performance of the models across 15 seeds.

\subsection{Results}
\subsubsection{Quantitative Results}
In Table \ref{tab:align_inputs}, we first compare how feature-aligned diffusion performs when the intermediate features are aligned to: a) expert features computed on the noise-free original training samples, and b) expert features computed on the noise-added training samples. We see a significant improvement when computing expert features on the noise-added inputs, making this the de-facto choice for our approach. We attribute this to the latent space of the downsampling block encoding information on the noisy image, as opposed to the noise-free image. That is, output at that stage will still contain the noise, before it is denoised through the upsampling path.

Figure \ref{fig:box} compares the expert classification accuracy on the synthetically generated images, between the baseline diffusion model and the feature-aligned diffusion model across different seeds, for the two fine-tuning pipelines. The expert model serves as a ``proxy'' for whether the generated images contain the features necessary to distinguish the respective classes. The dotted green line in Figure \ref{fig:box} shows the mean generation accuracy. \textbf{\ul{Feature-aligned diffusion \textit{consistently} outperforms the baseline diffusion approach} (by $9\%$ on average, across all seed settings). Feature-aligned diffusion improves model performances both with typical/vanilla fine-tuning and fine-tuning with DreamBooth, highlighting its \ul{synergistic potential with existing pipelines}}. 

For individual classes, we show the confusion matrices of the expert predictions on synthetic generations of the feature-aligned and baseline diffusion in Figures \ref{fig:cf_diff}, \ref{fig:base_cf_mat}. We compare this to the confusion matrix of the expert model predictions on the original data in Figure \ref{fig:cf_exp}. With synthetic generations, we see that the expert model tends to mis-classify ``tumor'', ``stroma'' and ``lympho'' classes as ``debris''. We see several more ``debris'' mis-classifications with the baseline diffusion. However, in Figure \ref{fig:cf_exp} we see that similar mis-classifications rarely occur on the original dataset. This implies that the diffusion model likely introduces specific artefacts that cause mis-classifications into the ``debris'' category. Such ``hallucinations'' occur in the context of diffusion models \cite{aithal2024understanding}, possibly related to the high intra-class variability of debris images.

For quantitatively measuring sample diversity, we follow prior work \cite{montoya2024mam} and measure SSIM (Structural Similarity Index Measure) which is used to assess the generation diversity of generative models. We compute SSIM between the generated synthetic images to measure the similarity between the generations -- \textit{lower} the SSIM, the \textit{better} the generation diversity. Figure \ref{fig:ssim} shows the SSIM values for the baseline and feature-aligned diffusion models for each class, and the overall averaged SSIM values across all classes. We see that \textbf{\ul{feature-aligned diffusion generations have better overall sample diversity compared to the baseline approach}}. In some cases the SSIM of feature-aligned diffusion is close to the original dataset indicating similar diversity as the original data. While there are a few classes where the baseline diffusion has a slightly lower SSIM, we note that \textit{SSIM alone does not imply ``accurate'' generations}. For instance, while baseline diffusion has a slightly lower SSIM for the ``mucosa'' class than feature-aligned diffusion, the corresponding classification accuracy is poor (Figure \ref{fig:base_cf_mat}) when compared to the feature-aligned model (Figure \ref{fig:cf_diff}) . Hence, putting the SSIM metric alongside the classification accuracy, shows that feature-aligned diffusion clearly outperforms the baseline diffusion approach.


\begin{figure}
\centering
  \includegraphics[width=0.39\textwidth]{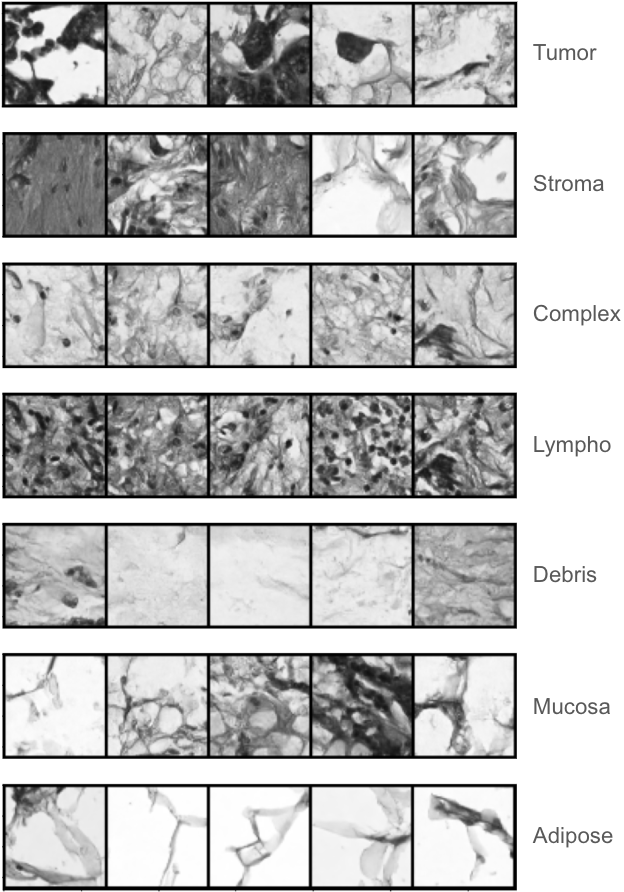}
  \caption{Example generations of baseline diffusion.}
  \label{fig:baseline}
\end{figure}

\subsubsection{Qualitative Results}
Synthetic generations of the feature-aligned diffusion model are shown in Figure \ref{fig:generations}. We see similarities between the synthetic generations, and images sampled from the original dataset. Some classes like ``debris'' exhibit higher intra-class variability in the images, whereas classes like ``adipose'' and ``mucosa'' show much less variability. Visually comparing the ``tumor'' and ``stroma'' class generations, we see that some properties of the generated images align with visual features of the ``debris'' class, possibly reinforcing the confusion matrix observations.

The synthetic generations of the baseline diffusion model is shown in Figure \ref{fig:baseline}. Comparing Figures \ref{fig:generations} and \ref{fig:baseline}, we see that particularly for certain classes like ``tumor'', ``complex'', ``debris'', and ``mucosa'', the synthetic generations of the baseline model are visually quite different from the original dataset. \textbf{\ul{Qualitatively, we see that the synthetic generations of the feature-aligned diffusion model match the original dataset more closely than the baseline diffusion model}}.




\section{Limitations and Future Work}
We explored feature-aligned diffusion, that is easily integrated into existing diffusion pipelines to improve synthetic generations. Our future work seeks to improve sample diversity further across all the classes for feature-aligned diffusion, through studying the influence of $w_1$ and $w_2$ in the loss terms. We will expand to additional datasets and explore the use of the synthetic data to improve the expert, alongside a deeper investigation of the source of the ``debris'' mis-classifications.


\subsection{Potential Applications to Other Domains}
Although we discuss feature-aligned diffusion specifically in the context of medical image synthesis, the approach may be applicable to other image synthesis domains, e.g., natural images (CIFAR10/100, ImageNet). The benefit of our approach may be more evident in cases where finer details need to be captured well in the synthetic generations (e.g., shapes of cells), and our future work seeks to quantitatively evaluate this hypothesis with feature-aligned diffusion for other domains.

\bibliographystyle{IEEEtran}
\bibliography{main}

\end{document}